\documentclass[lettersize,journal]{IEEEtran}

\usepackage{amsmath,amsfonts,amssymb,mathtools}

\usepackage{algorithmic}
\usepackage{algorithm}

\usepackage{array}
\usepackage[caption=false,font=normalsize,labelfont=sf,textfont=sf]{subfig}
\usepackage{stfloats}
\usepackage{multirow}
\usepackage{booktabs}

\usepackage{textcomp}
\usepackage{url}
\usepackage{verbatim}
\usepackage{graphicx}
\usepackage{xcolor}
\usepackage{ulem}

\usepackage{cite}
\usepackage{hyperref}
\usepackage[capitalize,noabbrev]{cleveref}

\usepackage{authblk}

\usepackage{amsthm}
\theoremstyle{plain}

\theoremstyle{definition}

\theoremstyle{remark}

\usepackage{pifont}
\usepackage{mdframed}
\usepackage{listings}
\begin{document}

\title{Improving Sample Efficiency of Reinforcement Learning with Background Knowledge from Large Language Models}

\author{Fuxiang Zhang, Junyou Li, Yi-Chen Li, Zongzhang Zhang, Yang Yu, Deheng Ye
\thanks{Fuxiang Zhang, Yi-Chen Li, Zongzhang Zhang, and Yang Yu are with National Key Laboratory for Novel Software Technology, Nanjing University, Nanjing 210023, China and School of Artificial Intelligence, Nanjing University, Nanjing 210023, China (e-mail: zhangfx@lamda.nju.edu.cn; liyc@lamda.nju.edu.cn; zzzhang@nju.edu.cn; yuy@nju.edu.cn).}
\thanks{Junyou Li and Deheng Ye are with Tencent, Shenzhen 518054, China (e-mail: junyouli@tencent.com; dericye@tencent.com).}

}

\markboth{Journal of \LaTeX\ Class Files}
{Shell \MakeLowercase{\textit{et al.}}: A Sample Article Using IEEEtran.cls for IEEE Journals}


\maketitle

\begin{abstract}
Low sample efficiency is an enduring challenge of reinforcement learning (RL). With the advent of versatile large language models (LLMs), recent works impart common-sense knowledge to accelerate policy learning for RL processes. However, we note that such guidance is often tailored for one specific task but loses generalizability. In this paper, we introduce a framework that harnesses LLMs to extract \textit{background knowledge} of an environment, which contains general understandings of the entire environment, making various downstream RL tasks benefit from one-time knowledge representation. We ground LLMs by feeding a few pre-collected experiences and requesting them to delineate background knowledge of the environment. Afterward, we represent the output knowledge as potential functions for potential-based reward shaping, which has a good property for maintaining policy optimality from task rewards. We instantiate three variants to prompt LLMs for background knowledge, including writing code, annotating preferences, and assigning goals. Our experiments show that these methods achieve significant sample efficiency improvements in a spectrum of downstream tasks from \textit{Minigrid} and \textit{Crafter} domains. 
\end{abstract}


\section{Introduction}
\begin{figure*}
\begin{center}
\includegraphics[width=\linewidth]{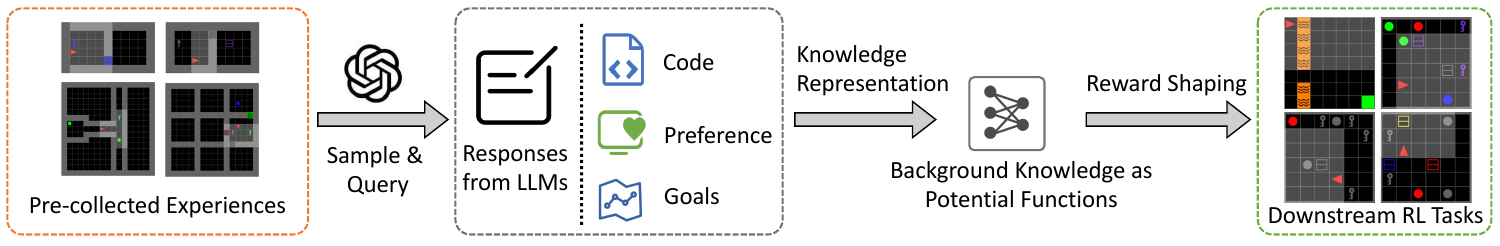}
\vskip -0.1in
\caption{An illustration of our framework to extract background knowledge from LLMs for reward shaping in downstream RL tasks. We sample experiences from pre-collected data and request LLM feedback in different forms including code, preference, or goals. The obtained feedback is represented as potential functions for potential-based reward shaping in downstream RL tasks. }
\label{fig:intro}
\end{center}
\vskip -0.2in
\end{figure*}

\IEEEPARstart{R}{einforcement} learning (RL)~\cite{sutton2018reinforcement} has achieved notable success in various domains including game AI~\cite{dqn, alphago, ye2020towards}, robotics~\cite{akkaya2019solving}, and natural language processing~\cite{instructgpt}. Conventionally, the success of RL hinges on extensive interactions, making low sample efficiency a huge challenge of RL literature \cite{sample-efficient-rl, exploration-survey}. This issue is particularly challenging in environments with sparse rewards, inspiring endeavors on designing auxiliary rewards and enhancing exploration. Researchers commonly involve various mechanisms to improve the sample efficiency of RL, such as intrinsic motivations \cite{imrl} based on novelty, curiosity, or uncertainty \cite{count-based, curiosity, rnd, noveld} and external knowledge sources including human annotations, knowledge bases, or foundational models \cite{language-reward-shaping, knowledge-exploration, foundation-rl}. 

As highlighted in recent research \cite{llm-survey, sparks-agi}, large language models (LLMs) such as GPT-4 \cite{gpt-4} have shown remarkable ability in instruction following with common-sense knowledge. This capability leads to innovative uses of LLMs in the field of RL, where recent studies develop approaches such as goal decomposition \cite{palo2023towards, ellm, zhao2023large} and writing codes \cite{eureka, dm-reward, text2reward}. Different from exploration with general human-concluded metrics like curiosity and uncertainty, LLM-assisted RL approaches provide training signals tailored for specific RL tasks more effectively. However, this specificity also comes with a downside. The heavy reliance on task-specific prompting may lead to an inability to produce reusable knowledge, particularly in open-ended domains where solving each task with exclusive prompting processes can be both costly and time-consuming.

In this paper, we utilize LLMs to represent the \textit{background knowledge} of an environment, thereby offering a more general and effective means to guide RL without repetitive LLM calls for different tasks. The proposed background knowledge, which is irrelevant to specific tasks, serves as the preliminary knowledge of an environment. For instance, the knowledge to avoid walls and obstacles in a grid world or to grab food in a survival game is generally useful to related environments regardless of executed tasks. We believe that the background knowledge can also be expressed as reward signals and propose a framework to extract and reuse such knowledge for downstream RL tasks based on designed desiderata of interaction-free, task-agnostic, and optimality-invariant requirements. 

As depicted in Figure~\ref{fig:intro}, we leverage a dataset containing experiences from multiple tasks to ground LLMs for decision-making problems, which avoids comprehensive interactions with the environment. Afterward, we particularly prompt the LLM to provide feedback on data samples based on its general understanding of the environment, forming task-agnostic background knowledge. The obtained knowledge is represented as potential functions for potential-based reward shaping \cite{reward-shaping}, a way that shapes RL processes without changing policy optimality, to accelerate RL in downstream tasks. As different potential functions may influence RL to different extents, we adopt three different variants for background knowledge representation inspired by previous research on harnessing LLMs, including writing code, annotating preference, and suggesting goals. Experimental results in the \textit{Minigrid} and \textit{Crafter} domains show that these variants all yield great sample-efficiency improvement. Furthermore, we discover the possibilities of reusing background knowledge for emerging task types or increasing task scales, proving the generalization ability of extracted background knowledge. We also include discussions on the sensitivity of our proposed variants with different choices of language models and data. Our contributions can be summarized as follows: 
\begin{itemize}
    \item We propose a framework that harnesses LLMs to provide background knowledge of an environment and thereby accelerates downstream RL tasks via potential-based reward shaping. 
    \item Based on the proposed framework, we develop three variants to represent background knowledge from LLM feedback for the reward-shaping procedure. 
    \item We show that acquired background knowledge can significantly improve sample efficiency and well generalize to previously unseen tasks. 
\end{itemize}

\section{Related Work}

\textbf{Reward Shaping in RL}. RL commonly faces poor sample efficiency \cite{sample-efficient-rl} especially in sparse reward tasks. Researchers establish theories and methods to enhance sample efficiency by enhancing agent exploration and exploitation \cite{exploration-survey, hindsight}. The most common approach to improving sample efficiency is reshaping the training signals of RL processes, i.e., reward shaping \cite{reward-shaping}. Recently, there has been a surge in shaping rewards with languages by utilizing human annotations \cite{language-reward-shaping, narration-reward-shaping}, deducing state novelty from language abstractions \cite{intrinsic-exploration, semantic-exploration}, or setting lingual goals \cite{ella, eager}. Unlike these methods, our work aims to extract the underlying background knowledge of the environment, which only adopts languages as an interface to acquire knowledge rather than exploiting language structures for RL. Although some prior works also try to integrate external knowledge for sample-efficient RL \cite{knowledge-exploration, ili}, they commonly adopt underlying structures of tasks such as symbolic input. In contrast, our work does not posit specific task designs but proposes a general framework to harness general-purpose LLMs for reward shaping.


\textbf{LLM-assisted RL}. Incorporated with proper techniques such as in-context learning~\cite{gpt-3} and chain-of-thought prompting~\cite{cot}, recent works show that LLMs are knowledgeable enough to master decision-making tasks \cite{voyager, spring}. However, directly using LLMs to solve complicated tasks can be difficult, whereas leveraging their powerful capabilities to guide RL processes can be a remedy. Some prior works directly use pretrained language models as policies and fine-tune their parameters with RL \cite{saycan, lid, glam}. Though effective, these methods require text-based states and actions for making decisions and usually face the problem of high computational cost. Another line of research tries to guide RL with the common-sense knowledge of LLMs, which is the domain our framework falls into. Some prior works utilize LLMs as an intermediate to provide language instructions \cite{llm4rl, lagr-seq, ellm} and train language-conditioned policies \cite{language-rl-survey}. Other works may focus on prompting LLMs to write code of reward functions \cite{dm-reward, eureka, text2reward} or check the completion of task goals \cite{reward-design-with-languages}. However, it is worth noting that these approaches are often tailored for specific RL tasks, where the particular mechanism may both invoke difficulties in applicability and bring considerate costs when deploying to different tasks. In our paper, we propose a general framework to utilize background knowledge for reward shaping, thereby improving sample efficiency for various downstream RL tasks within an environment. Different from previous works on harnessing LLMs for policy pretraining \cite{ellm}, our framework provides a lightweight reward-shaping process to integrate the knowledge of LLMs without querying them during RL processes. 


\section{Framework}

\subsection{Problem Statement}

We consider sequential decision-making problems in an open environment $\mathcal{E}$, in which a specific task $\mathcal{T}$ corresponds to a reinforcement learning (RL) problem. An agent interacts with the environment in discrete time to realize specific requirements defined by the task. Typically, an RL task can be modeled as a Markov decision process (MDP) ~\cite{sutton2018reinforcement} $\mathcal{T} = \left( \mathcal{S}, \mathcal{A}, r, P, \rho, \gamma \right)$. An agent starting at initial state $s_0 \sim \rho(\cdot)$ selects action $a_t \in \mathcal{A}$ and leads to a transition to $s_{t+1} \sim P(\cdot \mid s_t, a_t)$ with acquiring reward $r_t = r(s_t, a_t)$. An RL algorithm usually learns a policy $\pi(a\mid s)$ by maximizing the discounted cumulative reward, a.k.a. return, $R=\sum_{t=0}^\infty \gamma^t r_t$. Since an agent often perceives a view of the open environment, the received input in our experiments may be a partial observation. With a little misuse of notation, we do not explicitly distinguish states and observations in our methodology for simplicity since we do not delve into the theoretical properties of partial observability. A practical approach is to use the trajectory of agent history $\tau_t = (s_0, a_0, \dots, s_{t-1}, a_{t-1}, s_t)$ to represent agent state. Our work focuses on training policies $\pi(a\mid s)$ in any task $\mathcal{T}$ from an environment, where the task variance commonly comes from different targets and maps in our experiments. 

To tackle the problem of poor sample efficiency in RL, potential-based reward shaping \cite{reward-shaping, reward-shaping-2} is a useful technique for altering RL processes. Typically, this technique adds an additional reward term $F(s, s')$ to the original environmental reward $r$, whose value is computed with a defined potential function $\phi(s)$: 
\begin{equation}
    \label{eq:pbrs}
    F(s, s') = \gamma \phi(s') - \phi(s),
\end{equation}
where $s$ and $s'$ are the current state and the next state respectively and $\gamma$ is the discount factor of the MDP. An advantage of applying potential-based reward shaping is that it can maintain the policy optimality of original problems \cite{reward-shaping}. Consider an MDP $\mathcal{T}$ where $r_t$ is the extrinsic reward received from the environment at the time step $t$. We can denote $G = \sum_{t=0}^\infty r_t$ as the return of an episode in the original MDP. The concept of reward shaping refers to adding additional shaped reward $F(s, s')$ to the extrinsic reward $r$: 
\begin{equation}
    r' = r + F(s, s').
\end{equation}
Note that reward shaping is not to modify the reward function of the environment but to supplement additional rewards for computation. The purpose of the shaping function $F$ is often to provide heuristic domain knowledge to the problem when the agent transitions from state $s$ to $s'$. We thus define the return of an episode by 
\begin{equation}
    G' = \sum_{t=0}^{\infty} \gamma^t r' = \sum_{t=0}^{\infty} \gamma^t (r_t + F(s_t, s_{t+1})). 
\end{equation}
Potential-based reward shaping is a particular type of reward shaping. For potential-based reward shaping, the shaping function is in the form of 
\begin{equation}
    F(s, s') = \gamma \phi(s') - \phi(s),
\end{equation}
where $\gamma$ is the exact discount factor from the MDP. We call $\phi$ the potential function and $\phi(s)$ the potential of state $s$. Therefore, we can define the potential function $\phi: \mathcal{S} \rightarrow \mathbb{R}$ instead of defining $F: \mathcal{S} \times \mathcal{S} \rightarrow \mathbb{R}$. The reason for choosing this form of reward shaping is that it can converge to the optimal policy. Consider the return for one episode: 
\begin{equation}
\begin{aligned}
    G' &= \sum_{t=0}^{\infty} \gamma^t (r_t + F(s_t, s_{t+1})) \\
    &= \sum_{t=0}^{\infty} \gamma^t (r_t + \gamma \phi(s_{t+1} - \phi(s_t)) \\
    &= \sum_{t=0}^{\infty} \gamma^t r_t + \sum_{t=0}^{\infty} \gamma^{t+1} \phi(s_{t+1}) - \sum_{t=0}^{\infty} \gamma^t \phi(s_t)\\
    &= \sum_{t=0}^{\infty} \gamma^t r_t + \sum_{t=1}^{\infty} \gamma^{t} \phi(s_{t}) - \phi(s_0) - \sum_{t=1}^{\infty} \gamma^t \phi(s_t)\\
    &= G - \phi(s_0),
\end{aligned}
\end{equation}
where we can decompose $G'$ into the cumulative reward on the original MDP and the potential of the initial state $\phi(s_0)$. As the initial state $s_0$ can be any arbitrary state, we can easily extend this equality to a shaped Q-function, for example, $Q'(s, a) = Q(s, a) - \phi(s)$. Therefore, any RL algorithm that maximizes the cumulative reward or the Q-values will derive the same optimal policies after reward shaping, since the term $\phi(s)$ is not related to action selection. 

The optimality invariance of reward shaping is found in early research \cite{reward-shaping, reward-shaping-2} and widely applied in recent RL works \cite{language-reward-shaping, foundation-rl}. We note that although potential-based reward shaping can provide guarantees of the final results, it may alter the optimization dynamics in a way that either accelerates or slows down policy learning. A well-designed potential function $\phi(s)$ can decrease the time to convergence while bad ones may increase the time. In our work, we propose three approaches to design potential functions that can reflect the background knowledge of the environment from LLM feedback. We prove that our designs of the potential function can significantly improve sample efficiency, leading to a shorter convergence time. 

To ground LLMs for decision-making tasks, it is conventional to transform trajectories into text captions. Though this limitation may be addressed by further improvement on multimodal LLMs, we posit the existence of text captions in our pre-collected data for querying LLMs. Typically, the description of a trajectory can be derived from state captions and action names.


\begin{figure*}
\begin{center}
\includegraphics[width=\linewidth]{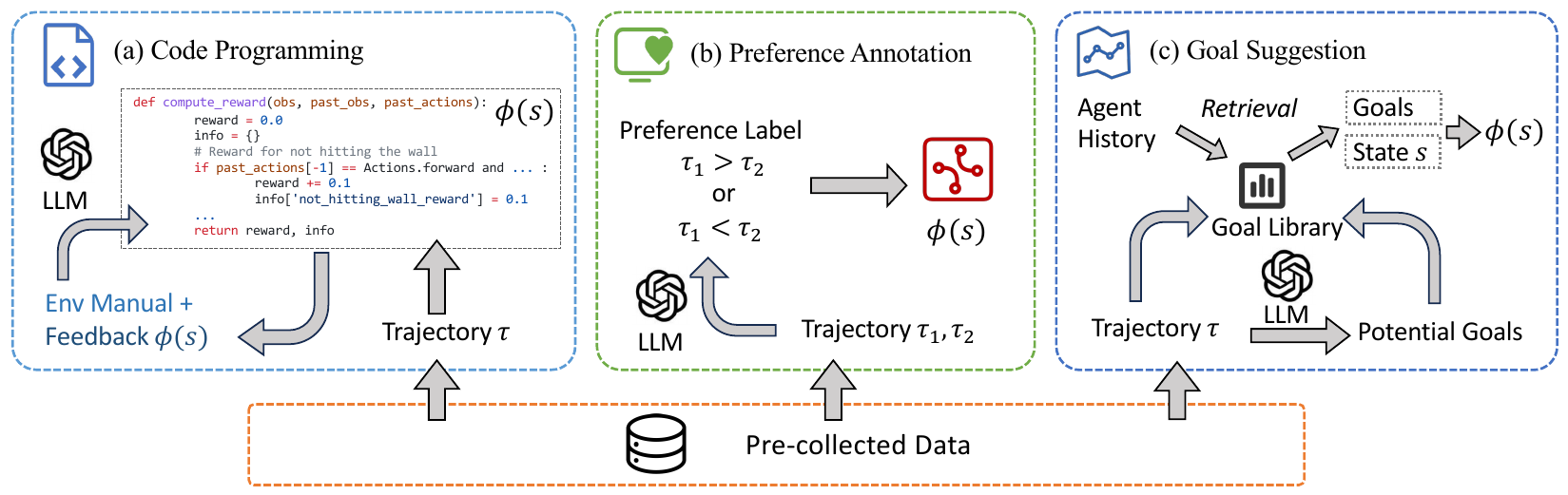}
\vskip -0.1in
\caption{The proposed three variants of background knowledge representation from pre-collected data. (a) We query an LLM to write code that returns high values for behaviors with desired background knowledge. We ask the LLM to iteratively improve the code from sampled results. (b) We prompt an LLM to annotate its preference over two trajectories and then learn the potential function $\phi(s)$ that decomposes preferences. (c) We sample trajectories from the dataset and ask the LLM to suggest potential goals. The pair of captions and goals are stored in a text-based goal library. To use the goal library for downstream RL, we retrieve results whose trajectories are similar to agent history and compute goal similarity with the current state. }
\label{fig:method}
\end{center}
\vskip -0.2in
\end{figure*}

\subsection{Reward Shaping with Background Knowledge}

In this section, we describe our framework that harnesses LLMs to obtain background knowledge of an environment, and further use it for accelerating downstream RL tasks. As there exist various ways to guide RL with LLM feedback, we summarize the desiderata of our framework below to better illustrate our requirements: 

\begin{itemize}
    \item \textbf{Interaction-free}. The process of acquiring background knowledge from LLMs should not be attached to an online RL process, which will involve a considerable amount of interactions with the environment. 
    \item \textbf{Task-agnostic}. The background knowledge acquired in our framework should represent a general understanding of the environment, rather than involving behaviors about specific tasks. 
    \item \textbf{Optimality-invariant}. We expect to not change policy optimality after injecting background knowledge into RL. We provide background knowledge for improving sample efficiency.
\end{itemize}

Based on these three principles, we design a framework to extract background knowledge for reward shaping as previously shown in Figure~\ref{fig:intro}. To satisfy the desiderata, we highlight three key procedures in the framework:

\textbf{Data collection}. To follow the interaction-free property, we pre-collect a dataset $\mathcal{D}=\{(s, a)\}$ by directly deploying an RL algorithm in an open environment without specifying its task goal. Therefore, the dataset only contains states and actions to delineate possible interactions in the environment. Specifically, we adopt RND \cite{rnd} and save its interactive data periodically, following the standard in offline RL literature \cite{d4rl}. As an intrinsically motivated RL approach, RND tends to explore novel states in the open environment. Therefore, the collected data can contain diverse and useful behaviors for concluding background knowledge. 


\textbf{Background knowledge representation}. To ground LLMs for the environment, we iteratively sample trajectories from the pre-collected data and ask LLMs for their feedback. During this procedure, the LLM makes assessments based on captions of agent trajectories but keeps unknown about the task goal, producing task-agnostic background knowledge. The knowledge is then represented as a potential function $\phi(s)$ by gathering LLM feedback over data samples, where we defer different variants for background knowledge representation to the next section. 

\textbf{RL with reward shaping}. Using the potential function $\phi(s)$, we can adopt potential-based reward shaping for downstream RL tasks according to Equation~\eqref{eq:pbrs}, which shapes RL processes with preserving policy optimality. We adopt the PPO algorithm \cite{ppo} based on the implementation from CleanRL \cite{cleanrl} to train policies in different tasks.

The above three phases achieve our proposed desiderata, composing a common framework to extract and reuse background knowledge for downstream RL tasks. 
We note that the implementation of data collection and the RL process can be realized by different approaches. Specifically, we use the PPO algorithm \cite{ppo} with the CleanRL \cite{cleanrl} implementation, a stable and succinct PPO version for adopting our reward-shaping technique. The only modification to the training process is to add the auxiliary reward to the original environmental reward. To adapt to our selected environments, we use convolutional neural networks to extract features from the input observation. Then the features are fed into different multi-layer perceptrons (MLPs) to compute the action logits or value functions. Due to the different observation spaces of our evaluated environments, the feature processing networks are slightly different.  We show the network structure of RL agents and also report the hyperparameters of the algorithms in our appendix, where most of the hyperparameters are directly taken from the CleanRL implementation despite some coefficients are related to environment sampling and auxiliary rewards. 



\begin{table*}
\caption{Comparisons of the proposed three variants on background knowledge representations from LLM feedback. }
\label{table:variants}
\centering
\begin{tabular}{c|cccc}
\toprule
Variant & Sources of background knowledge & \begin{tabular}[c]{@{}c@{}}Additional model \\during RL\end{tabular} & \begin{tabular}[c]{@{}c@{}}Text captions\\during RL\end{tabular} & \begin{tabular}[c]{@{}c@{}}Unstructured input\\support (e.g., images)\end{tabular} \\
\midrule
\textsc{BK-Code} & LLM-coded functions & \ding{56} & \ding{56} & \ding{56} \\
\textsc{BK-Pref} & LLM's preference from data & parameterized $\phi(s)$ & \ding{56} & \ding{52} \\
\textsc{BK-Goal} & goals suggested by LLMs & a sentence encoder & \ding{52} & \ding{52} \\
\bottomrule
\end{tabular}
\end{table*}

\section{Background Knowledge Representation} 

In this section, we introduce three different variants to request LLMs for background knowledge. As previous works have made several endeavors to ground LLMs for decision-making tasks, we find that relevant approaches on prompting LLMs are also effective in representing background knowledge and propose three variants including code programming (Section~\ref{section:coding}), preference annotation (Section~\ref{section:preference}), and goal suggestion (Section~\ref{section:goal}). We design effective yet simple approaches to query LLMs for background knowledge and transform results into required potential functions. 

\subsection{Code Programming (\textsc{BK-Code})}
\label{section:coding}

With pretraining in code data, LLMs exhibit powerful abilities in programming and reasoning \cite{codex, codegen, llm-code-survey}, propelling successful applications to write code for guiding RL \cite{dm-reward, text2reward, eureka}. Most of these works propose to prompt an LLM to write code of reward functions to depict desired targets. As coding is a direct approach to conveying LLM knowledge, we thus introduce an iterative prompting procedure to harness LLMs to program and improve their written code. We design straightforward prompts to ask LLMs to provide a Python function whose arguments come from the current observation and agent histories. By providing agent trajectories, LLMs can analyze agent behaviors based on more thorough considerations. Our prompt contains three portions: 

\begin{itemize}
    \item \textbf{Environmental information}. We attach the environment description and necessary constants and variables of the environment, serving as a header of the code. 
    \item \textbf{Trajectory samples}. We sample trajectory data from the pre-collected data and caption them to better ground LLMs for the environment.
    \item \textbf{Feedback from samples}. We use the current code to compute results from sampled data and append the results as an evaluation. 
\end{itemize}

To be more specific, we prompt LLMs to write code for a function \texttt{def compute\_reward(obs, past\_obs, past\_actions)}. Although we use the computation results as the potential values $\phi(s)$, we name the function name \texttt{compute\_reward} with a little misuse, since it is better for LLMs to understand. The parameters of the function include the current observation, historical observations, and historical actions, where we truncate the length of the historical sequence to $5$ for efficiency consideration. The LLM is supposed to write a function that returns the expected value of the potential function and also a dictionary that describes different portions of the result. This additional dictionary gives the LLM diverse feedback and thus helps improve the code. This helpful information will also be attached to the next query. There may be situations where the code provided by the LLM cannot pass the Python interpreter, especially for weak LLMs. In this case, we attach the error log reported by the interpreter and ask the LLM to fix potential issues. The LLM will have multiple retry chances to write a runnable code. Otherwise, we will stop the iteration and use the latest successful code as the final result. For each iteration, we sample trajectories from the dataset and feed them into the written code for the output values and the information dictionary. Afterward, we attach the text caption of the trajectories along with the dictionary information and ask the LLM to improve the code. We repeat this process for $20$ rounds, which is a moderate number since we find that involving additional iterations may result in too much unnecessary or inaccurate code.

\subsection{Preference Annotation (\textsc{BK-Pref})}
\label{section:preference}

As LLMs show the potential to be a helpful evaluation tool \cite{llm-eval, llm-eval-2}, many prior works utilize the feedback of LLMs to align learned models \cite{constitutional-ai, rlaif, motif}. We note that the preference over trajectories can also be a useful metric to delineate background knowledge and thus develop an approach that learns the potential function from LLM feedback. We sample two trajectories $\tau^0, \tau^1$ from pre-collected data and ask LLMs to provide preferences over the given pair. The sampled trajectory sequences are truncated to $H$ steps $\tau = (s_1, a_1, \dots, s_H, a_H)$ for fair comparisons. The preference label can be $y=0$ for $\tau^0 > \tau^1$ or $y=1$ for $\tau^1 > \tau^0$, resulting in a preference dataset $\mathcal{D}_{\text{pref}} = \{(\tau^0, \tau^1, y)\}$. Following previous preference-based RL studies \cite{pbrl1, pbrl2}, we can define a preference predictor following the Bradley-Terry model \cite{bt-model}:
\begin{equation}
    \label{eq:bt-model}
    P[\tau^1 > \tau^0] = \frac{\exp\left( \sum_t \phi(s_t^1) \right)}{\exp ( \sum_t \phi(s_t^0)) + \exp ( \sum_t \phi(s_t^1) )},
\end{equation}
where $\phi(s)$ is the potential function. Therefore, we can learn $\phi(s)$ by maximizing the likelihood of Equation~\eqref{eq:bt-model} in our annotated dataset $\mathcal{D}_\text{pref}$ in the following objective:
\begin{equation}
\label{eq:reward-learning}
\begin{aligned}
    \mathcal{L}(\theta) = - \underset{(\tau^0, \tau^1, y) \in \mathcal{D}_{\text{pref}}}{\mathbb{E}} 
    &\big[ y\log P[\tau^1 > \tau^0; \phi] \\
    &+ (1-y) \log P[\tau^0 > \tau^1; \phi] \big],
\end{aligned}
\end{equation}
where $\theta$ is the parameters of $\phi$. Following \cite{preference-transformer}, we adopt the Transformer structure to compute $\phi(s)$ with historical information. For each preference label, we sample two trajectories $\tau^0, \tau^1$ from pre-collected data and ask LLMs to provide preferences over the given pair. The trajectory sequence is truncated to a length of $H=5$ and is then captioned to compose the input prompt. We annotate labels including $y=0$ for $\tau^0 > \tau^1$ and $y=1$ for $\tau^1 > \tau^0$, forming a dataset $\mathcal{D}$ of labeled data. In addition, we also find that LLMs may refuse to provide a rank in the process because they equally preferred the provided two trajectories. To utilize this portion of data, we add an additional log-likelihood term to the original loss on the unlabeled data $\mathcal{D}'$:
\begin{equation}
\label{eq:smoothing}
    \mathcal{L}'(\theta) = - \underset{(\tau^0, \tau^1) \in \mathcal{D}'}{\mathbb{E}} 
    \big[ \log P[\tau^1 > \tau^0; \phi] + \log P[\tau^0 > \tau^1; \phi] \big].
\end{equation}
We mix $\mathcal{L}(\theta)$ and $\mathcal{L}'(\theta)$ when optimizing the preference predictor. To parameterize the potential function $\phi(s)$, it is conventional to use a neural network with the input of agent state $s$. However, we find that it can be more helpful to make $\phi$ a non-Markovian function according to the previous work Preference Transformer \cite{preference-transformer} since the previous observations and actions contain useful information to judge the potential of the current state. Therefore, we adopt the Transformer structure to capture sequential information. We feed the input sequence $(s_1, a_1, \dots, s_H, a_H)$ into the Transformer and aggregate the output representations on each state $s_t$ to compute the logits with an additional preference attention layer proposed by \cite{preference-transformer}. The embedding dimension of the Transformer is $128$ with $1$ layer and $1$ header. We adopt the same feature extractor of the RL policy to process input observation and a simple embedding layer to process the action. 

\subsection{Goal Suggestion (\textsc{BK-Goal})}
\label{section:goal}

Another way to involve background knowledge is to guide the agent with potential goals of the environment. Recent works \cite{ellm, llm4rl, sama} propose several techniques to query LLMs for helpful goals and thus guide the agent to visit states of interest. In our approach, we try to discover such potential goals in an offline manner. We first caption sampled trajectories $\tau$ from the dataset and ask LLMs for potential goals the agent can reach. Then we store the output goals $g$ along with the trajectory caption, forming a text-based goal library for further utilization. During the RL process, we introduce a retrieval procedure to select appropriate goals from the goal library according to the agent history. When the agent is in state $s_t$, we use the text form of agent history $\tau_{t_1}$ to retrieve an exhaustive goal list $g_1, \dots, g_k$ from the top-$K$ most similar trajectories. The similarity metrics are calculated using embeddings provided by a pretrained language model as the sentence encoder \cite{sentence-transformer}. Then we compute the cosine similarity $\sigma(s_t, g_i)$ between the current trajectory $\tau_t$ and the goal $g_i$ as the achievement of each goal. 
The value of the potential function in $s$ is the maximal similarity of potential goals $\phi(s) = \max_{i} \sigma(s, g_i)$ since it provides the degree of how well the agent completes the most potential goal. 

To harness LLMs for goal suggestion, we also caption the trajectory sequence and add the caption to the prompt. We then ask the LLM to provide potential goals based on the history of the agent. We ask the LLMs to list all possible goals in the form of an unordered list and stone all the goals along with the trajectory caption. The pair of trajectory captions and corresponding goals compose our desired goal library. For downstream RL, we retrieve the top-$K$ similar pairs according to the similarity between the query text and the text caption from the library, where $K$ is set to $3$. For our approach, we deploy a naive method by computing the cosine similarity for each text caption, but we note that approaches to accelerating this retrieval process are well-studied, which goes beyond our scope. We adopt a small-scale pretrained BERT model \cite{bert} from HuggingFace\footnote{\url{https://huggingface.co/prajjwal1/bert-small}} as our sentence encoder. 

\begin{figure*}
\vskip 0.2in
\begin{center}
\includegraphics[width=\linewidth]{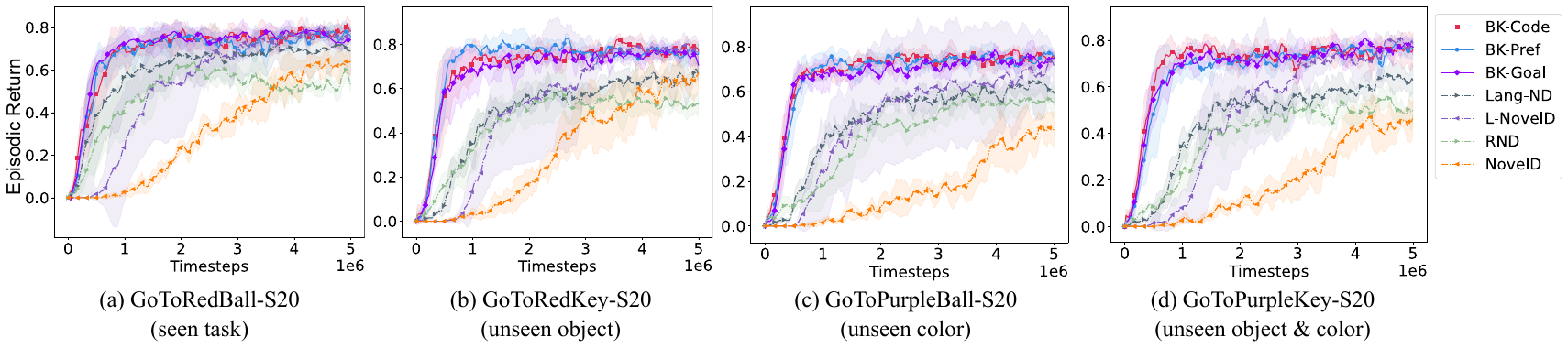}
\vskip -0.1in
\caption{Average episodic returns of compared methods in different BabyAI \textit{goto} tasks of the Minigrid environment. Task goals containing the color \textit{purple} and the object type \textit{key} do not appear in the collected datasets. }
\label{fig:minigrid-main}
\end{center}
\end{figure*}

\begin{figure*}
\begin{center}
\includegraphics[width=\linewidth]{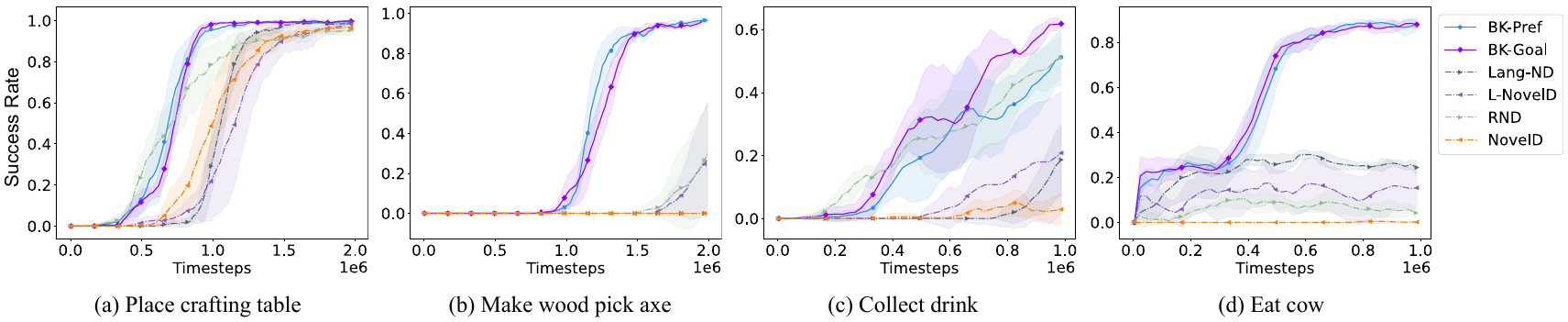}
\vskip -0.1in
\caption{Average success rates of compared methods in different downstream tasks of the Crafter environment. For each task, the agent only acquires a reward when completing the corresponding achievement. }
\label{fig:crafter-main}
\end{center}
\vskip -0.2in
\end{figure*}


We summarize the properties of our proposed three variants on background knowledge representation in Table~\ref{table:variants}. Although these methods all leverage LLMs to express background knowledge, the differences in representing potential functions result in different behaviors during downstream RL tasks. \textsc{BK-Code} does not require an additional parameterized model but calculates auxiliary rewards solely using written code. In contrast, \textsc{BK-Pref} and \textsc{BK-Goal} require pretrained models to express potential functions. The similarity computation in \textsc{BK-Goal} requires text-form environmental descriptions, limiting its applications to environments that can provide a text captioner. The other two methods only require text captions in collected data samples, making it practical to only caption data samples rather than implementing a general captioner. Besides, \textsc{BK-Code} is infeasible for RL tasks with unstructured input such as images and texts since we cannot directly parse such input using code. 

\textbf{Principles on prompt design}. The prompts in background knowledge representation typically contain two portions, the environmental information and the mission description. An exception comes from \textsc{BK-Code}, which requires additional code snippets for programming. We make the designed prompts modularized and succinct, applicable to different environments with moderate efforts in replacing the environment-related prompt. In our supplementary materials, we list all used prompts, discuss how to adapt to other domains, and provide some example responses from different environments and LLMs.

\section{Experiments}

In this section, we aim to evaluate the effectiveness of using background knowledge in downstream RL tasks from the following aspects\footnote{Code available at \url{https://github.com/mansicer/background-knowledge-rl}}: (1) \textbf{Sample efficiency improvement}. Is reward shaping with background knowledge more efficient than prior sample-efficient RL approaches? (2) \textbf{Generalizability of background knowledge}. When the data is from a subset of tasks, can the derived background knowledge help a broader range of tasks in the environment? (3) \textbf{Sensitivity of background knowledge}. We analyze two key factors that affect the qualities of knowledge -- the language model and data size -- to evaluate the sensitivity of the proposed three variants. We conduct experiments in two popular environments: (1) \textit{Minigrid} \cite{minigrid}, an environment with easily configurable grid-world tasks frequently used to evaluate exploration methods, and (2) \textit{Crafter} \cite{crafter}, an open-ended environment where agents should collect achievements while discovering survival strategies. As stated in Table~\ref{table:variants}, only \textsc{BK-Goal} from our variants require text captions during downstream RL training. The other two methods only require captions on the pre-collected data, making it possible to caption the dataset for environments that may be difficult to write a rule-based captioner. Although previous research \cite{ellm} also tries to train captioners based on vision language models, here we simply adopt existing captioning functions from prior works. For Minigrid, we adopt the text captioner from a previous work GLAM \cite{glam} to caption states. For the Crafter environment, we use the text captioner from the SmartPlay benchmark \cite{smartplay}. We directly borrow captioners from existing works to avoid the effectiveness and reproducibility issues of designing a specific captioner for our framework.

\begin{figure}
\begin{center}
\includegraphics[width=0.8\linewidth]{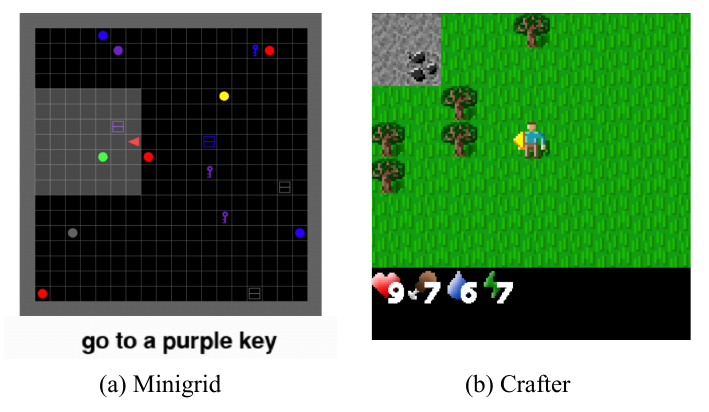}
\caption{Rendered game frames from two used environments: (a) \textit{Minigrid} and (b) \textit{Crafter}.}
\label{fig:app-env}
\end{center}
\end{figure}

\begin{figure}
\begin{center}
\includegraphics[width=0.95\linewidth]{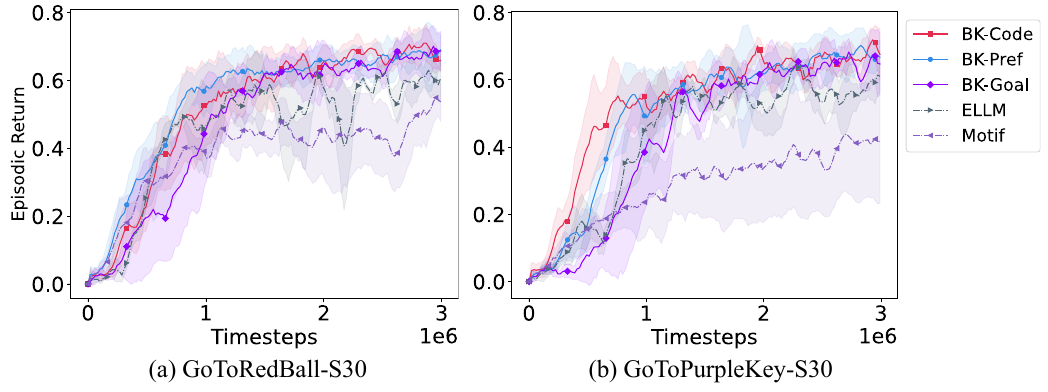}
\caption{Average episodic returns of our methods compared with ELLM and Motif in different BabyAI \textit{goto} tasks of the Minigrid environment.}
\label{fig:ellm-motif}
\end{center}
\end{figure}


We introduce baselines, including classic exploration approaches and sample-efficient RL methods using language abstraction. We adopt RND \cite{rnd} and NovelD \cite{noveld} as popular exploration baselines that generally show better sample efficiency than PPO. We also introduce Lang-ND \cite{semantic-exploration} and L-NovelD \cite{intrinsic-exploration} as methods that enhance sample efficiency by providing intrinsic rewards through languages. Similar to ours, these methods do not require specific task information but usually train language models to provide features during the RL process. However, our approaches either require no text input or only use pretrained sentence encoders for inference, which is a more efficient way. 
Our reported results are averaged over $5$ runs with error bars denoting the standard deviation. 

\begin{figure*}
\begin{center}
\includegraphics[width=\linewidth]{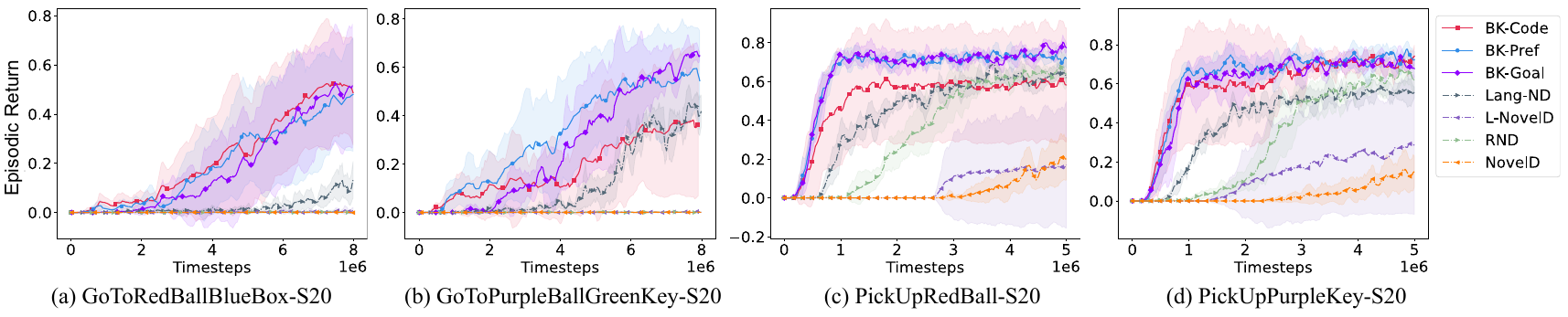}
\vskip -0.1in
\caption{Average episodic returns of compared methods in different emerging BabyAI tasks of the Minigrid environment including (a, b) \textit{goto-seq} tasks and (c, d) \textit{pickup} tasks. }
\label{fig:minigrid-extend}
\end{center}
\vskip -0.2in
\end{figure*}

\subsection{Performance on Downstream Tasks}

\textbf{Minigrid} \cite{minigrid} contains a spectrum of configurable tasks in a grid environment. We adopt the \textit{Minigrid} environment \cite{minigrid} (Figure~\ref{fig:app-env}(a)) from its official GitHub repository\footnote{\url{https://github.com/Farama-Foundation/Minigrid}}. For this environment, we mainly use its \textit{goto} and \textit{pickup} tasks types from the \textit{BabyAI} domain \cite{babyai} to build tasks. To specify the domain knowledge for a comprehensive analysis, we mainly focus on the \textit{goto} tasks from the BabyAI series \cite{babyai}, which is a subset of original Minigrid tasks. A \textit{goto} task typically requires the agent to navigate to a specific object in a map with multiple distractors. The target object type can be a ball, a box, or a key with a specific color from red, blue, green, or purple. Similarly, a \textit{pickup} task requires the agent navigate to an object and perform a pickup action. The original environment registration in the code repository is text-conditioned, where the agent may receive different text instructions for separate episodes. To unify the task goal, we make specifications for the environment, making it generate consistent goals during RL processes. We create a series of \textit{goto} and \textit{pickup} tasks for our experiments. A \textit{goto} task typically requires the agent to navigate to a specific kind of object, e.g., a red ball in the grid world. As for its extension, the \textit{goto-seq} task requires the agent to sequentially navigate to two different objects in one episode, bringing more difficulties for agent exploration. A \textit{pickup} task additionally asks the agent to perform a \textit{pickup} action after navigating to the object. The original BabyAI environments usually contain map sizes ranging from $5$ to $8$, which can be simple for exploration methods. To create more challenging benchmarks, we scale the map size to $20-30$ to conduct our experiments. To introduce diverse downstream tasks for evaluation, we design different kinds of unseen tasks on purpose. The data for background knowledge representation is collected from tasks without targets of the object type \textit{key} and the color \textit{purple}. However, during the downstream RL training, we aim to train policies in these tasks within one-time knowledge representation. Therefore, we can evaluate the effectiveness of acquired background knowledge in a spectrum of seen and unseen tasks. 

In Figure~\ref{fig:minigrid-main}, we show the average episodic returns of different methods in four downstream tasks with different targets. We find that our proposed three variants generally outperform the compared baselines in all tasks. Notably, when conducting experiments in tasks with unseen object types and colors, the improvement of sample efficiency of our methods is still obvious. The results indicate that our way of representing background knowledge does acquire task-agnostic background knowledge of the environment, thus accelerating policy learning in unseen tasks. We also find that Lang-ND and L-NovelD, which both adopt state captions to provide intrinsic motivation, perform better than classic exploration baselines, demonstrating that leveraging text features can be useful to improve sample efficiency. 

\textbf{Crafter} \cite{crafter} is a 2D survival game (Figure~\ref{fig:app-env}(b)) drawing inspiration from the popular Minecraft game. The environment is procedurally generated and partially observable, where the agent needs to complete achievements such as collecting and crafting. The original Crafter game rewards agents based on the accomplishment of achievements and health changes. Based on the achievement list, we create downstream tasks corresponding to each achievement where the agent can only get a reward when completing it. We also adopt the environment from its official repository\footnote{\url{https://github.com/danijar/crafter}}. However, the original action space in Crafter merges different operations like eating, collecting, and attacking into a single `do' action. Though successfully reducing the action space for exploration, this implementation may hinder understanding of environmental logic. Following the solution in \cite{ellm}, we split this action into several actions including `eat', `drink', `attack', and `collect', making it more aligned to actual agent behaviors. The transformation on the action space also helps explain actions during text captioning. We create different downstream tasks in the Crafter environment by splitting the achievement list of the environment into different tasks. We evaluate our proposed methods in these tasks except for \textsc{BK-Code} since the observation space in Crafter is image-based. As shown in Figure~\ref{fig:crafter-main}, we find that the two applicable approaches, \textsc{BK-Pref} and \textsc{BK-Goal}, still exhibit superior performance in most downstream tasks. For crafting tasks that require specific action sequences like making a wood pick axe, introducing background knowledge can significantly improve sample efficiency. 

Some recent works also consider learning RL policies with LLM knowledge besides our compared baselines, such as Motif \cite{motif} and ELLM \cite{ellm}. However, these methods are not directly comparable to our method due to different training paradigms and the inability of prompt design in each task. Motif also takes the idea of learning from LLM-labeled preference, also known as RL from artificial intelligence feedback (RLAIF) \cite{constitutional-ai, rlaif, rlaif-pjc}, which directly uses the preference data to augment rewards. Unlike Motif, our framework focuses on more general background knowledge of a domain to avoid complex prompt design and costly LLM queries. In addition, ELLM adopts a different paradigm to generate possible goals during policy pretraining. In contrast, our framework does not require the engagement of LLMs within the interactions with the LLMs but extracts background knowledge from LLMs in an interaction-free manner. Although these methods cannot realize an efficient paradigm like ours, we can compare them in each single task. In Figure~\ref{fig:ellm-motif}, we show the results in a map with size 30 to enlarge its difficulty, where these baselines extract knowledge from each single task data while our methods still use previous acquired knowledge. We find that our three methods can still perfrom well compared to these baselines, indicating the effectiveness of integrating LLM knowledge via reward shaping.


\begin{figure}
\begin{center}
\includegraphics[width=\linewidth]{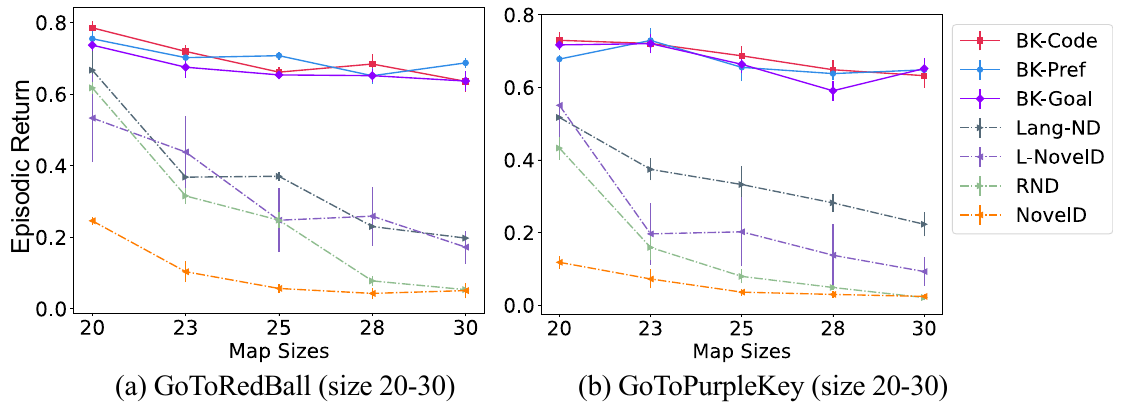}
\vskip -0.1in
\caption{Average episodic return at the $2$M-th time step in \textit{goto} tasks with different map sizes ranging from $20$ to $30$. }
\label{fig:minigrid-scaling}
\end{center}
\vskip -0.2in
\end{figure}

\subsection{Generalizability of Background Knowledge}

In this section, we further test the generalizability of acquired background knowledge, aiming to find whether our framework of using background knowledge can accelerate policy learning for more distinct tasks. We conduct experiments in tasks with different task types and map sizes from the highly configurable Minigrid environment to evaluate the performance of our methods. 

\textbf{Effectiveness on emerging task types}. The data used for acquiring background knowledge contains experiences from the \textit{goto} tasks, which only require the agent to navigate to one object. Here we further introduce two additional task types to examine whether the background knowledge can provide more general training signals. The extra task types include (1) the \textit{goto-seq} task, where the agent should navigate to two distinct objects sequentially, and (2) the \textit{pickup} task, where the agent needs to execute a ``pickup'' action after successfully navigating to the target. As shown in Figure~\ref{fig:minigrid-extend}, our methods still exhibit superior performance compared to other baselines. Specifically, the \textit{goto-seq} task type is relatively difficult as it requires exhaustive exploration to find two desired objects. Our compared baselines can hardly solve this kind of task while our methods can solve the problem without previously encountering these tasks. We find that \textsc{BK-Code} is less stable and shows less promising performance than the other two variants in some tasks, indicating that this form of knowledge representation may not be as generalizable as other variants. 

\textbf{Scaling to larger map sizes}. When the map becomes larger, the agent needs to execute more actions precisely to acquire the reward, resulting in increasing difficulty in exploring and exploiting such reward signals. To this end, we configure a series of \textit{goto} tasks with different map sizes to examine whether reward shaping with background knowledge can scale well to larger maps. We evaluate our methods in this task series and plot the average episodic returns at the $2$M-th time step in Figure~\ref{fig:minigrid-scaling}. Notably, our proposed methods can all maintain high sample efficiency with increasing task difficulty. In contrast, the compared baselines mostly exhibit significant performance drops when the map becomes larger. The results indicate that our three variants can scale to larger maps well. 

\begin{figure}
\begin{center}
\includegraphics[width=\linewidth]{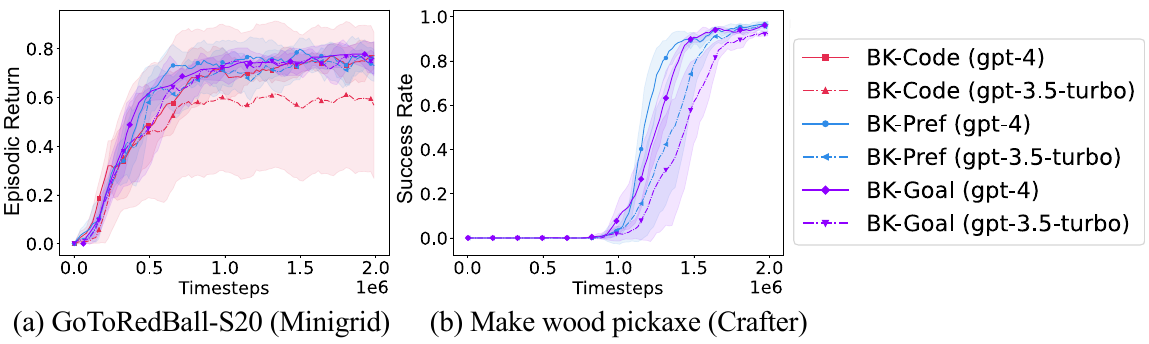}
\vskip -0.1in
\caption{The performance of proposed methods using different GPT-series models for background knowledge representation.}
\label{fig:ablation}
\end{center}
\vskip -0.2in
\end{figure}

\begin{figure}
\begin{center}
\includegraphics[width=\linewidth]{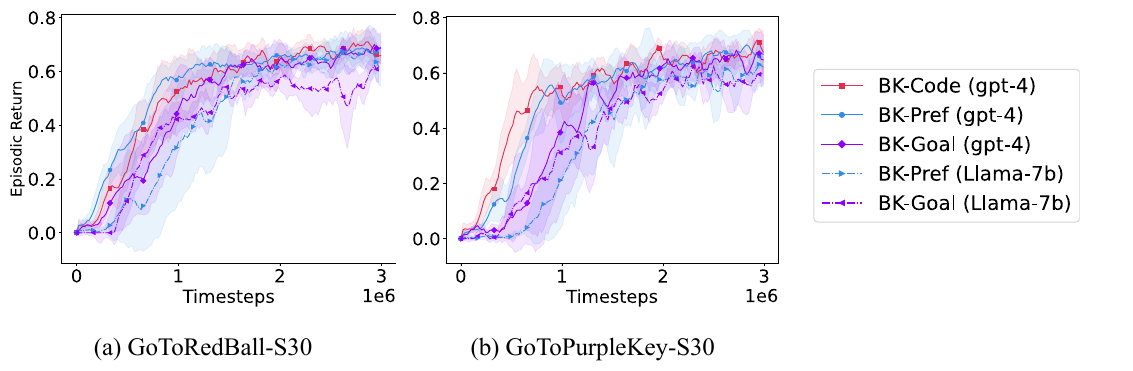}
\vskip -0.1in
\caption{The performance of proposed methods using GPT-4 and Llama-2 language models for background knowledge representation.}
\label{fig:ablation-llama}
\end{center}
\vskip -0.2in
\end{figure}

\subsection{Sensitivity of Background Knowledge}

In this section, we conduct experiments to find out the sensitivity of learned background knowledge under the conditions of different language models and data. In Figure~\ref{fig:ablation}, we evaluate our three variants with two LLMs, \texttt{gpt-3.5-turbo} and \texttt{gpt-4}, separately. The results show that \textsc{BK-Code} is the most sensitive approach when we turn to a weaker LLM. Since the quality of code strongly correlates to the capability of LLMs, we also find that weaker LLMs have a higher probability of generating code with runtime errors. The performance drop when using \texttt{gpt-3.5-turbo} in \textsc{BK-Pref} and \textsc{BK-Goal} is hardly observed in Minigrid but is more evident in the crafting task from Crafter, where an LLM with lower capability may have a more superficial environment understanding. To further investigate the effectiveness of our framework under much weaker open-source language models, we also test the performance of \textsc{BK-Pref} and \textsc{BK-Goal} when using a 7B version of Llama-2 chat model in Figure~\ref{fig:ablation-llama}. We omit the \textsc{BK-Code} method with the Llama version since we find that the Llama-2 chat model is not capable to generate valid code in our case. We find that \textsc{BK-Pref} and \textsc{BK-Goal} using Llama models perform slightly worse than the methods with GPT models, indicating that the capability of LLMs may affect the quality of extracted knowledge. However, thanks to our effective knowledge representations from preferences or goals, the performance loss is not significant and these algorithms can still solve the task with knowledge from a weak Llama-2 7B model.

\begin{figure}
\begin{center}
\includegraphics[width=\linewidth]{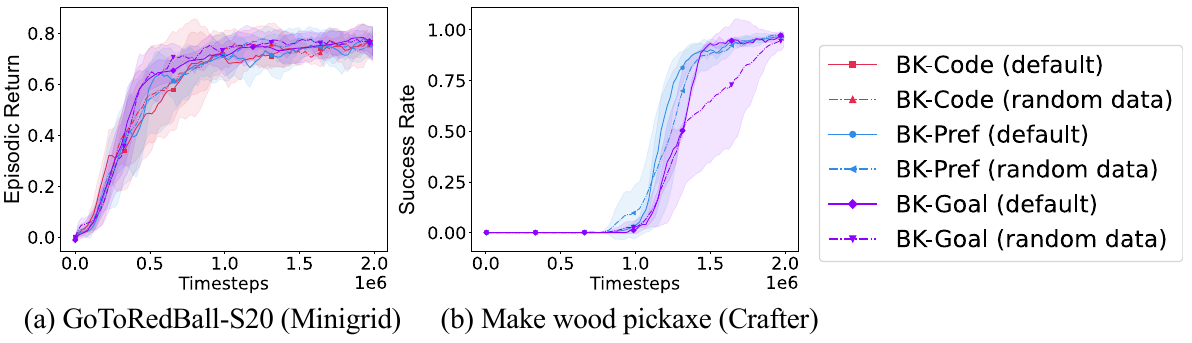}
\vskip -0.1in
\caption{The performance of proposed methods using different data sources for background knowledge representation.}
\label{fig:dataset}
\end{center}
\vskip -0.2in
\end{figure}

Besides, we try to discover the connection between used data and RL performance, as LLMs may not be able to derive sufficient knowledge from low-quality data. we present the policy performance when using data collected by random policies for background knowledge representation, abbreviated as the \textit{random data} approach. As shown in Figure~\ref{fig:dataset}, we still observe significant sample efficiency improvements for these variants. For Minigrid, we find that background knowledge derived from random data is sufficient to accelerate policy learning in \textit{goto} tasks. For the more difficult task of making a wood pickaxe in Crafter, random policies may not be able to provide useful trajectories, limiting the knowledge LLMs can provide. 

\section{Conclusions and Limitations}

In this paper, we propose a novel framework to extract and reuse background knowledge of an environment by harnessing LLMs. Leveraging a pre-collected dataset, we design three variants, \textsc{BK-Code}, \textsc{BK-Pref}, and \textsc{BK-Goal}, to represent background knowledge from LLM feedback as helpful potential functions. With a unified reward-shaping framework, we adopt the derived potential functions for different downstream tasks. Our experiments in two different environments show that our framework significantly improves sample efficiency in downstream RL tasks and the LLM-concluded knowledge can even generalize to unseen tasks beyond the scope of provided data. We also present a detailed analysis of the generalizability and sensitivity of our framework under different conditions. 

Our work has a few limitations. First, we did not optimize the prompting mechanisms for the proposed variants. It is important to note that our current prompting methods for LLMs may not be the most effective, and future research could explore more sophisticated approaches. Additionally, although we employ a modular and concise prompt design, creating a prompt with environmental information for new domains still requires a moderate effort. Future studies that aim to simplify and automate these procedures could be highly beneficial.

\section*{Acknowledgments}
This work is supported by the National Science Foundation of China (62276126) and the
Tencent AI Lab (RBFR2023011).

{\appendix[]

\begin{table*}
\caption{Typical time cost of all compared methods. }
\label{table:time-cost}
\centering
\begin{tabular}{l|lllllll}
\toprule
Method    & BK-Code & BK-Pref & BK-Goal & RND    & NovelD & Lang-ND & L-NovelD \\
\midrule
Time cost & 6.57h   & 6.67h   & 25.42h  & 13.17h & 11.51h & 14.68h  & 17.87h   \\
\bottomrule
\end{tabular}
\end{table*}

\begin{figure}[h!]
\begin{center}
\includegraphics[width=\linewidth]{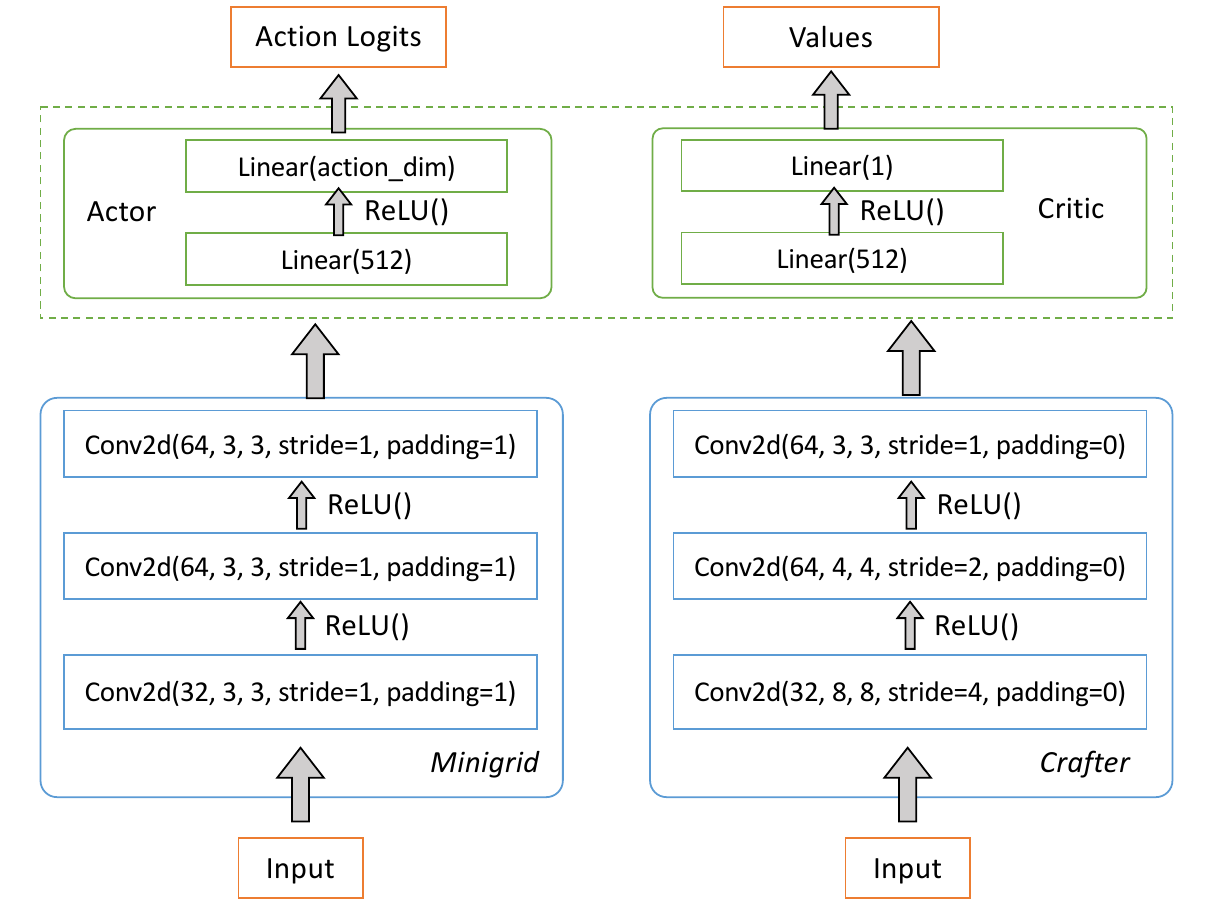}
\caption{The network structures of RL agents trained for the Minigrid and Crafter tasks.}
\label{fig:app-network}
\end{center}
\end{figure}

\begin{table}[h!]
\caption{Hyperparameters for downstream RL training}
\label{table:hyperparameters}
\centering
\begin{tabular}{ll}
\toprule
\multicolumn{1}{c}{Hyperparameter} & \multicolumn{1}{c}{Value} \\ 
\midrule
Discount factor $\gamma$ & 0.99                            \\
Learning rate                       & 0.0003                          \\
Extrinsic reward coefficient        & 10.0                            \\
GAE factor $\lambda$ & 0.95                            \\
Parallel workers                    & 8 (Minigrid), 16 (Crafter)      \\
Batch size                          & 1024 (Minigrid), 4096 (Crafter) \\
Importance sampling clipping        & 0.1                             \\
Entropy loss coefficient            & 0.01                            \\
Value loss coefficient              & 0.5                             \\
Gradient clipping factor            & 0.5                             \\
\bottomrule
\end{tabular}
\end{table}

\begin{figure}
\begin{center}
\includegraphics[width=\linewidth]{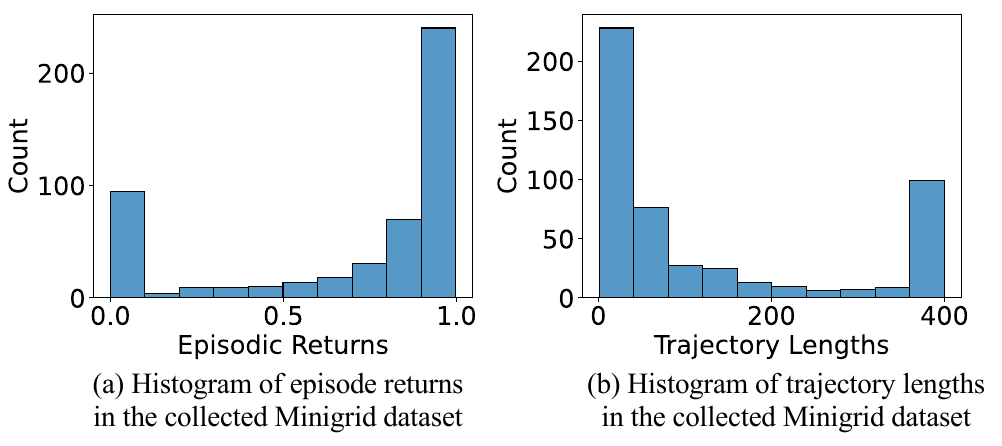}
\caption{The histograms of episodic returns and trajectory lengths in the used Minigrid dataset.}
\label{fig:data-properties}
\end{center}
\end{figure}

\begin{table}[t]
\caption{Typical LLM query cost of BK-Code, BK-Pref, and BK-Goal. }
\label{table:llm-call}
\centering
\begin{tabular}{l|lll}
\toprule
Method  & \# LLM calls & Cost (gpt-3.5-turbo) & Cost (gpt-4) \\
\midrule
BK-Code & 50           & \$0.15               & \$18         \\
BK-Pref & 2000         & \$1.5                & \$90         \\
BK-Goal & 500          & \$0.9                & \$54         \\
\bottomrule
\end{tabular}
\end{table}

\subsection{Details of Data Collection}

As stated in our main paper, we deploy an RND algorithm in the environment and collect data periodically. To be specific, we train RND for $5$M steps and evaluate the learned policy for each $500000$ step. During the evaluation process, we run the policy in $50$ episodes and save the data as our dataset. For Minigrid, we run the RND algorithm simultaneously in multiple tasks without fixing the task types, which means that the collected unlabeled trajectories may come from a distribution of tasks. We specify the task range by restricting the target types as mentioned in our experiments section, where the object type key and the color purple do not appear in the pre-collected data. For the Crafter environment, we directly deploy an RND algorithm in the original environment. The agent will try to maximize the raw Crafter reward based on completing achievements and keeping healthy. We present the properties of collected datasets in Figure~\ref{fig:data-properties}.

\subsection{Details of Downstream RL training}

We use the PPO algorithm \cite{ppo} with the CleanRL \cite{cleanrl} implementation, a stable and succinct PPO version for adopting our reward-shaping technique. The only modification to the training process is to add the auxiliary reward to the original environmental reward. To adapt to our selected environments, we use convolutional neural networks to extract features from the input observation. Then the features are fed into different multi-layer perceptrons (MLPs) to compute the action logits or value functions. We show the network structure of RL agents in Figure~\ref{fig:app-network}. Due to the different observation spaces of the two environments, the feature processing networks are slightly different. We also report the hyperparameters of the RL algorithms in Table~\ref{table:hyperparameters}, where most of the hyperparameters are directly taken from the CleanRL implementation despite some coefficients related to environment sampling and auxiliary rewards. 

\subsection{Cost of Our Methods}
Our method is less time-consuming since it only requires one-time knowledge extraction and does not call LLM during RL. We show the time cost of all methods in the \texttt{BabyAI-GoToRedBall-S30} task as an example. In Table~\ref{table:time-cost}, we find that our methods are time-efficient, except for BK-Goal which needs to compute text embeddings during training. Generally, the time cost of all our methods is comparable with these baselines. In addition, we also present typical values of LLM calls and costs for our three approaches in Table~\ref{table:llm-call}. Our framework is cost-efficient since it does not involve LLM calls during RL. In contrast, some prior works like ELLM typically requires millions of LLM calls since it needs interactions with LLMs in the RL pretraining process.

}

\bibliographystyle{IEEEtran}
\bibliography{reference}

\end{document}